\pdfoutput=1

\documentclass[11pt]{article}

\usepackage{ACL2023}
\usepackage{amsfonts}
\usepackage{amsmath}
\usepackage{amssymb}
\usepackage{bm}
\usepackage{booktabs}
\usepackage[T5,T1]{fontenc}
\usepackage{footnotehyper}
\usepackage{graphicx}
\usepackage[utf8]{inputenc}
\usepackage{multirow}
\usepackage{caption}
\usepackage{subcaption}
\usepackage{xspace}
\usepackage{svg}
\usepackage{tabularx}
\usepackage{times}
\usepackage{latexsym}

\usepackage{graphicx}
\usepackage{algorithm}
\usepackage{algpseudocode}
\usepackage{euflag}
\usepackage{arydshln}

\usepackage{microtype}
\usepackage{inconsolata}

\usepackage{anyfontsize}

\DeclareTextSymbolDefault{\ohorn}{T5}
\DeclareTextSymbolDefault{\uhorn}{T5}
\DeclareUnicodeCharacter{0300}{`}

\usepackage{todonotes}
\makeatletter
\newcommand*\iftodonotes{\if@todonotes@disabled\expandafter\@secondoftwo\else\expandafter\@firstoftwo\fi}
\makeatother

\definecolor{edolime}{rgb}{0.9,1,0.3}

\newcommand{\method}[1]{{\textsc{#1}}}
\newcommand{\loss}{\mathcal{L}}

\newcommand{\bistil}{{\textsc{BiStillation}}\xspace}

\newcommand{\rparagraph}[1]{\vspace{1.8mm}\noindent\textbf{#1.}}
\newcommand{\sparagraph}[1]{\vspace{0.0mm}\noindent\textbf{#1.}}

\title{Distilling Efficient Language-Specific Models for Cross-Lingual Transfer}

\author{Alan Ansell$^1$~~~Edoardo Maria Ponti$^{2,1}$~~~Anna Korhonen$^1$~~~Ivan Vuli\'{c}$^1$ \\
        $^1$Language Technology Lab, University of Cambridge\\
        $^2$University of Edinburgh\\
        \texttt{aja63@cam.ac.uk}\\
}

\begin{document}
\maketitle
\begin{abstract}
Massively multilingual Transformers (MMTs), such as mBERT and XLM-R, are widely used for cross-lingual transfer learning. While these are pretrained to represent hundreds of languages, end users of NLP systems are often interested only in individual languages. For such purposes, the MMTs' language coverage makes them unnecessarily expensive to deploy in terms of model size, inference time, energy, and hardware cost. We thus propose to extract compressed, language-specific models from MMTs which retain the capacity of the original MMTs for cross-lingual transfer. This is achieved by distilling the MMT \textit{bilingually}, i.e., using data from only the source and target language of interest. Specifically, we use a two-phase distillation approach, termed \bistil: (i) the first phase distils a general bilingual model from the MMT, while (ii) the second, task-specific phase sparsely fine-tunes the bilingual `student' model using a task-tuned variant of the original MMT as its `teacher'. We evaluate this distillation technique in zero-shot cross-lingual transfer across a number of standard cross-lingual benchmarks. The key results indicate that the distilled models exhibit minimal degradation in target language performance relative to the base MMT despite being significantly smaller and faster. Furthermore, we find that they outperform multilingually distilled models such as DistilmBERT and MiniLMv2 while having a very modest training budget in comparison, even on a per-language basis. We also show that bilingual models distilled from MMTs greatly outperform bilingual models trained from scratch. 

\end{abstract}

\section{Introduction}

\begin{figure}[!t]
    \centering
    \includegraphics[width=0.97\linewidth]{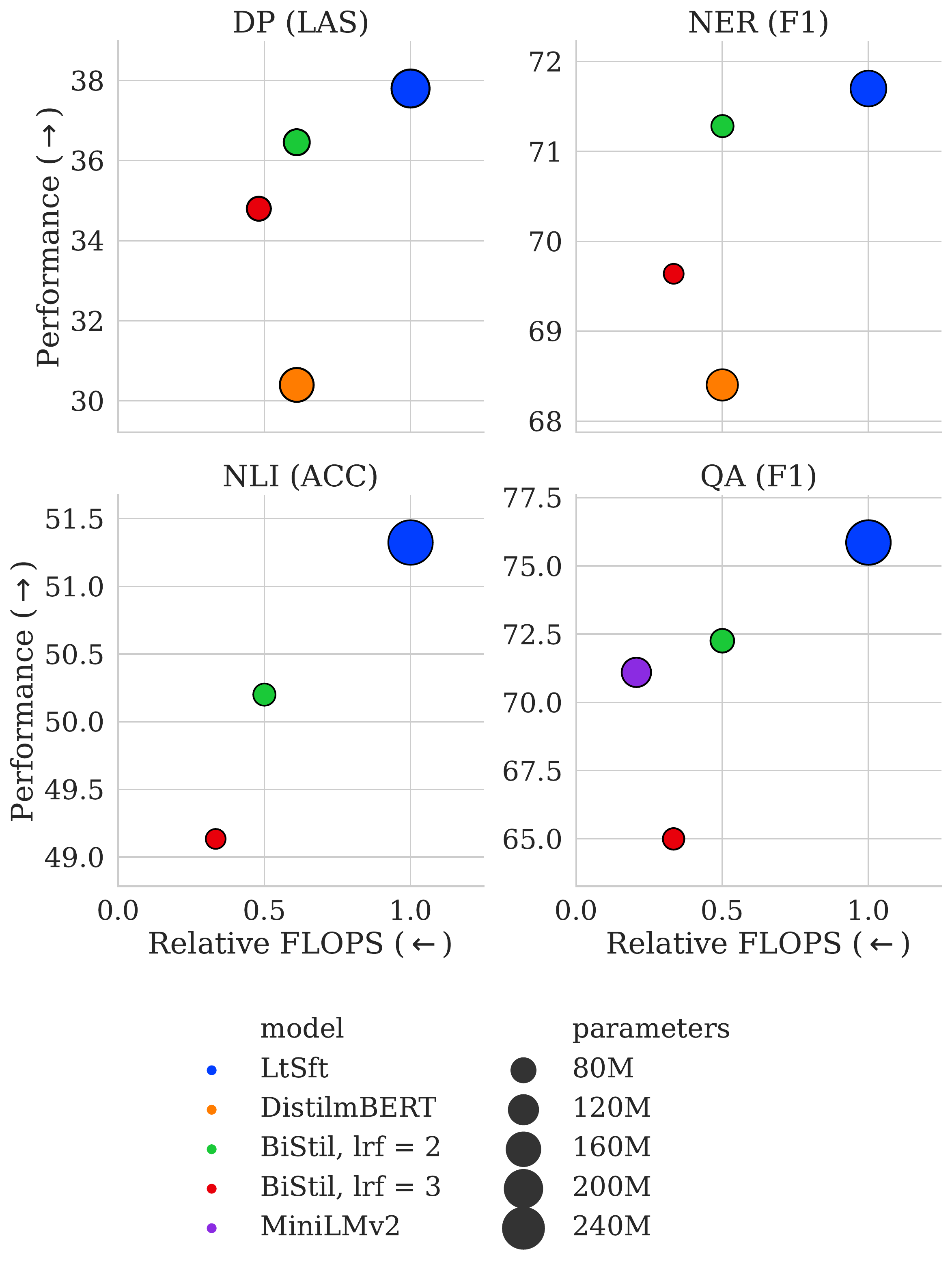}
    \caption{Tradeoff between parameter count, inference FLOPs and averaged performance for our \method{BiStil} models for cross-lingual transfer and several baselines.}
    \label{fig:efficiency-plots}
\end{figure}

Massively multilingual Transformers (MMTs), pretrained on unlabelled data from hundreds of languages, are a highly effective tool for cross-lingual transfer \citep{devlin-etal-2019-bert,conneau-etal-2020-unsupervised,chung-etal-2020-rethinking,he-etal-2021-debertav3}.
However, they suffer from several limitations as a result of their ample language coverage. Firstly, aiming to represent many languages within their parameter budget and dealing with the training signals from different languages might result in negative interference. This is known as the ``curse of multilinguality'' \citep{conneau-etal-2020-unsupervised}, which impairs the MMT's transfer capabilities \citep{pfeiffer-etal-2022-lifting}. Secondly, in practice
people are often interested in using or researching NLP systems in just a \textit{single} language.
This makes the MMTs \textit{unnecessarily costly} in terms of storage, memory, and compute and thus hard to deploy. This especially impacts communities which speak low-resource languages, which are more likely to have limited access to computational resources \citep{alabi-etal-2022-adapting}.

In this work, we address the question: \textit{can we increase the time-efficiency and space-efficiency of MMTs while retaining their performance in cross-lingual transfer?} Knowledge distillation \citep{hinton-etal-2015-distilling} is a family of general methods to achieve the first goal by producing smaller, faster models \citep[\textit{inter alia}]{sanh-etal-2019-distilbert,jiao-etal-2020-tinybert} and has also been applied to MMTs specifically. However, when the distilled MMT is required to cover the same number of languages as the original model, whose capacity is already thinly stretched over hundreds of languages, the ``curse of multilinguality'' asserts itself, resulting in a significant loss in performance \citep{sanh-etal-2019-distilbert}.

As a consequence, to achieve the best possible performance with reduced capacity, we depart from the practice of retaining all the languages from the original MMT in the distilled model. Instead, we argue, we should cover only \textit{two} languages, namely the source language and the target language of interest. In fact, distilling just \textit{one} language would fall short of the second goal stated above, namely facilitating cross-lingual transfer, as a monolingually distilled model would be unable to learn from a distinct source language during task-specific fine-tuning. Maintaining cross-lingual transfer capabilities, however, is crucial due to the paucity of labelled task data in many of the world's languages in most tasks \citep{ponti-etal-2019-modeling,joshi-etal-2020-state}.

In particular, we propose a method for \textit{bilingual distillation} of MMTs, termed \bistil, inspired by the two-phase recipe of \citet{jiao-etal-2020-tinybert}. We start from a \textit{``student''} model, initialized by discarding a subset of layers of the original \textit{``teacher''} MMT, as well as the irrelevant part of its vocabulary. In the first, \textit{``general''} phase of distillation, unlabelled data is used to align the the hidden representations and attention distributions of the student with those of the teacher. In the second, \textit{task-specific} phase, the student is fine-tuned for the task of interest through guidance from a task-adapted variant of the teacher. Rather than fully fine-tuning the student during this second phase, we instead use the parameter-efficient Lottery-Ticket Sparse Fine-Tuning (LT-SFT) method of \citet{ansell-etal-2022-composable}. Parameter-efficient task fine-tuning enables a system to support multiple tasks with the same distilled compact model, without unnecessarily creating full model copies per each task. 


We evaluate our efficient \textit{``bistilled''} models on a range of downstream tasks from several benchmarks for multilingual NLP, including dependency parsing from Universal Dependencies \citep[UD;][]{ud-2.7}, named entity recognition from MasakhaNER \citep{adelani-etal-2021-masakhaner}, natural language inference from AmericasNLI \citep{ebrahimi-etal-2022-americasnli}, and QA from XQuAD \citep{artetxe-etal-2020-cross}. We evaluate the model performance as well as its space efficiency (measured in terms of parameter count) and time efficiency (measured in terms of FLOPs and inference time). We compare it against highly relevant baselines: bilingual models pretrained from scratch and two existing multilingual distilled models, DistilmBERT \citep{sanh-etal-2019-distilbert} and MiniLMv2 \citep{wang-etal-2021-minilmv2}.

We find that while our bilingually distilled models are twice or thrice smaller and faster than the original MMT, their performance is only slightly degraded, as illustrated in Figure~\ref{fig:efficiency-plots}. Our method outperforms the baselines by sizable margins, showing the advantages of (i) bilingual as opposed to multilingual distillation, and (ii) distilling models from MMTs rather than training them from scratch. We hope that our endeavour will benefit end-users of multilingual models, and potential users under-served by currently available technologies, by making NLP systems more accessible. The code and models are publicly available at \url{https://github.com/AlanAnsell/bistil}.

\section{Background}

\subsection{Cross-Lingual Transfer with MMTs}

Prominent examples of MMTs include mBERT \citep{devlin-etal-2019-bert}, XLM-R \citep{conneau-etal-2020-unsupervised} and mDeBERTa \citep{he-etal-2021-debertav3}, among others.

\citet{pires-etal-2019-multilingual} and \citet{wu-dredze-2019-beto} showed that mBERT is surprisingly effective at \textit{zero-shot cross-lingual transfer}. Zero-shot cross-lingual transfer is a useful paradigm when there is little or no training data available for the task of interest in the target language, but there is training data available in some other \textit{source} language. In the simplest form of zero-shot cross-lingual transfer, the model is trained on source language data and is then used without modification for inference on target language data. While this generally works quite well for high-resource languages, transfer performance degrades for low-resource languages, especially those under-represented or fully unseen by the MMT during its pretraining \citep{lauscher-etal-2020-zero,pfeiffer-etal-2020-mad,ansell-etal-2021-mad-g,adelani-etal-2021-masakhaner,ebrahimi-etal-2022-americasnli}.

\subsection{Modular Adaptation of MMTs}
\label{ss:modular}
Because MMTs divide their capacity among many languages, they may often perform sub-optimally with respect to a single source or target language. Furthermore, we are sometimes interested in a target language not covered by the MMT. A naive solution to these problems is to prepare the MMT with continued pretraining on the target language before proceeding to task fine-tuning. While this can improve performance, \citet{pfeiffer-etal-2020-mad} show that a more effective approach is to perform this continued pretraining in a parameter-efficient manner, specifically with the use of \textit{adapters} \citep{rebuffi-etal-2017-learning,houlsby-etal-2019-parameter}. The resulting language-specific adapter is known as a \textit{language adapter}. When the task fine-tuning is also learned in the form of an adapter (\textit{task adapter}), \citeauthor{pfeiffer-etal-2020-mad} demonstrate that zero-shot transfer can be achieved by composing arbitrary language and task adapter pairs. 

\citet{ansell-etal-2022-composable} extend this idea to a new parameter-fine tuning method, \textit{sparse fine-tuning} (SFT). An SFT of a model is where only a sparse subset of its pre-trained parameters are fine-tuned, i.e. an SFT of a pretrained model $F$ with parameters $\bm{\theta}$ can be written as $F(\cdot\ ;\ \bm{\theta} + \bm{\phi})$, where the \textit{difference vector} $\bm{\phi}$ is sparse \cite{Sung:2021neurips}. Language and task SFTs with difference vectors $\bm{\phi}_L$ and $\bm{\phi}_T$ respectively are composed through addition, i.e. yielding $F(\cdot\ ;\ \bm{\theta} + \bm{\phi}_L + \bm{\phi}_T)$. SFTs are learned through a procedure called ``Lottery Ticket Sparse Fine-Tuning'' (LT-SFT), based on the Lottery Ticket algorithm of \citet{frankle-carbin-2019-lottery}. The $k$\% of parameters which undergo the greatest absolute change during an initial full fine-tuning phase are selected as tunable parameters during the second ``sparse'' phase which yields the final SFT. 

As SFT composition exhibited somewhat better zero-shot cross-lingual transfer performance across a range of tasks than adapter composition, and SFTs avoid the inference time slow-down incurred by adapters at inference time, we adopt this parameter-efficient approach throughout this work. However, we note that other modular and parameter-efficient architectures can also be tried in future work \cite{Pfeiffer:2023survey}.


\rparagraph{Multi-Source Training}
\label{sec:multi-source}
\citet{ansell-etal-2021-mad-g} show that multi-source task adapter training, where a task adapter is trained using data from several source languages simultaneously, yields large gains in cross-lingual transfer performance as a result of the task adapter learning more language-agnostic representations. \citet{ansell-etal-2022-composable} find similarly large gains from multi-source training of task SFTs. An important aspect of cross-lingual transfer with SFTs is that the source language SFT is applied during task SFT training. This requires each batch during multi-source training to consist of examples from a single source language, for which the relevant language SFT is applied during the corresponding training step.


\subsection{Distilling Pretrained Language Models}
Knowledge distillation \citep{bucilua-etal-2006-model,hinton-etal-2015-distilling} is a technique for compressing a pretrained large ``teacher'' model into a smaller ``student'' model by training the student to copy the behavior of the teacher. Whereas during standard pretraining, the model receives a single ``hard'' label per training example, during distillation the student benefits from the enriched signal provided by the full label distribution predicted by the teacher model. \citet{sanh-etal-2019-distilbert} use this technique to produce \textit{DistilBERT}, a distilled version of BERT$_\text{base}$ \citep{devlin-etal-2019-bert} with 6 instead of the original 12 layers, and \textit{DistilmBERT}, a corresponding distilled version of multilingual BERT. There has been extensive subsequent work on distillation of pretrained language models, but with less focus on distilling MMTs in particular. 


\section{\bistil: Methodology} 
\label{sec:methodology}

\sparagraph{Overview}
We are interested in providing NLP capabilities with limited computational resources in a specific target language $T$ which lacks training data in the tasks of interest. A common paradigm in previous work~\citep{pfeiffer-etal-2020-mad,ansell-etal-2022-composable} is to use cross-lingual transfer with an MMT in conjunction with parameter-efficient task and language adaptation to support multiple tasks without adding a large number of additional parameters per task, see \S\ref{ss:modular}. Our goal in this work is to replace the highly general MMT, plus optional language adaptation, with a target language-specific model which maintains the benefits of cross-lingual transfer. 

An obvious first attempt would be to simply distil the MMT into a smaller model using only text in the target language. However, this monolingual distillation approach is insufficient, as during task fine-tuning, the monolingually distilled student model no longer ``understands'' the source language. Indeed, our preliminary experiments confirmed the intuition that this approach is inadequate. This problem can be overcome through \textit{bilingual} distillation, where text from both the source and target language is used to train the student model.\footnote{This is similar to the idea of bilingual language adapters proposed by \citet{parovic-etal-2022-bad}, which obtain superior transfer performance by adapting the MMT to both source and target language simultaneously, removing the need to use different and possibly incompatible language adapters during training and inference.}


Therefore, our aim is to devise a method for deriving from an MMT $M$ a smaller model $M'_{S,T,\tau}$ to perform a given task $\tau$ in the target language $T$ given only training data in the source language $S$. Our approach is inspired by the two-stage distillation paradigm of \citet{jiao-etal-2020-tinybert}. In the first, ``general'' phase, a bilingual student model $M'_{S,T}$ is distilled from $M$ using the same unsupervised task (e.g., masked language modeling) that was used for $M$'s pretraining. In the second, ``task-specific'' phase, $M'_{S,T,\tau}$ is produced by fine-tuning $M'_{S,T}$ using $M_{\tau}$ as its teacher, where $M_{\tau}$ is derived from $M$ by fine-tuning it for task $\tau$. The following sections explain the details of these phases.


\subsection{Distillation Method} \label{sec:distillation-method}
Let $L_T$ be the number of Transformer layers in the teacher model, indexed from 1 to $L_T$. The number of student model layers $L_S$ is required to evenly divide $L_T$. We define the downscaling stride as $s = \frac{L_T}{L_S}$.

Following \citet{jiao-etal-2020-tinybert}, the loss functions of the two distillation phases make use of three components, (i) \textit{attention-based}, (ii) \textit{hidden state-based}, and (iii) \textit{prediction-based}. Attention-based loss is defined as follows:
\begin{align}
    \loss_\text{attn} = \frac{1}{L_S} \sum_{i=1}^{L_S} \texttt{MSE}(A_i^S, A_{i \cdot s}^T).
\end{align}
Here, $A_i^S$ and $A_i^T \in \mathbb{R}^{l \times l}$ refer to the attention distribution\footnote{Here, for ease of implementation within the Huggingface Transformers library \citep{wolf-etal-2020-transformers}, we differ from \citet{jiao-etal-2020-tinybert}, who use raw attention scores.} of Transformer layer $i$ of the student and teacher model, respectively; $l$ refers to the input sequence length; \texttt{MSE}() denotes mean squared error loss.

Hidden state-base loss is defined as follows:
\begin{align}
    \loss_\text{hidden} = \frac{1}{L_S + 1} \sum_{i=0}^{L_S} \texttt{MSE}(H_i^S, H_{i \cdot s}^T),
\end{align}
where $H_i^S$ and $H_i^T \in \mathbb{R}^{l \times d}$ refer to the hidden representations output by Transformer layer $i$ of the student and teacher model, respectively, or the output of the embedding layer when $i = 0$. Note that we assume that the student and teacher share the same hidden dimensionality $d$.

Finally, the prediction-based loss is defined as
\begin{align}
    \loss_{\text{pred}} = \texttt{CE}(\bm{z}^S, \bm{z}^T),
\end{align}
where $\bm{z}^S$ and $\bm{z}^T$ are the label distributions predicted by the student and teacher model, respectively, and \texttt{CE} denotes cross-entropy loss.

The intuition behind using attention-based and hidden state-based loss for our purposes is as follows. We (i) require good monolingual performance in the source and target language, but we also (ii) must preserve the existing alignment between these languages in the MMT which would consequently facilitate transfer between them. The intuition is that encouraging the student's intermediate representations to match those of the teacher will help to preserve this alignment.

We next describe how these loss components are employed in each phase of \bistil.

\subsection{Stage 1: General Bilingual Distillation} \label{sec:general-distillation}
\sparagraph{Initialization} 
We initialize all parameters of the student model by copying those of the teacher model, but retaining only the Transformer layers whose indices are multiples of $s$.

\rparagraph{Vocabulary Reduction} 
Our distilled models can dispose of the many irrelevant tokens in the base MMT's vocabulary, i.e. those which are not frequently used in either the source or target language of interest, an idea previously proposed by \citet{abdaoui-etal-2020-load}. During initialization, the vocabulary of the student model is selected by retaining only the tokens of the teacher's vocabulary whose unigram probability in either the source or target language corpus is $\ge 10^{-6}$.

\rparagraph{Teacher Language Adaptation} 
As we wish to be able to produce distilled models for languages not covered in the base MMT, and to obtain the best possible performance for languages which are covered, we employ language adaptation of the teacher MMT with language-specific SFTs \citep{ansell-etal-2022-composable} applied on top of the original MMT during distillation.\footnote{Put simply, additionally applying language-specific SFTs `skews' the MMT towards those particular languages.} Since it draws examples from two languages, each with its own language SFT, bilingual distillation becomes a special case of multi-source training as described in \S\ref{sec:multi-source}. At each training step, either the source or target language is selected at random with equal probability; the batch is composed of sequences drawn from the training corpus of the chosen language, and a pretrained SFT for that language is applied to the teacher MMT.

\rparagraph{Objective} 
The overall loss function for this phase is given by the sum of the attention-based and hidden state-based loss. Omitting the prediction-based loss here has the advantage of avoiding the need to evaluate the distribution of tokens predicted by the MLM head, which is costly because of the considerable size of MMTs' embedding matrices.

\begin{table*}[!t]
    \centering
    \def\arraystretch{0.999}
    \scriptsize
    \resizebox{\linewidth}{!}{\begin{tabular}{m{0.15\textwidth}  m{0.15\textwidth}  m{0.15\textwidth} m{0.08\textwidth}  m{0.35\textwidth}}
        \toprule
        \textbf{Task} & \textbf{Target Dataset} & \textbf{Source Dataset} & \textbf{MMT} & \textbf{Target Languages} \\
        \cmidrule(lr){2-5}
        Dependency Parsing (DP) & Universal Dependencies 2.7 \citep{ud-2.7}
        & Universal Dependencies 2.7 \citep{ud-2.7} & mBERT & Arabic$^\dagger$, Bambara, Buryat, Cantonese, Chinese$^\dagger$, Erzya, Faroese, Japanese$^\dagger$, Livvi, Maltese, Manx, North Sami, Komi Zyrian, Sanskrit, Upper Sorbian, Uyghur \\
         \cmidrule(lr){2-5}
        Named Entity Recognition (NER) & MasakhaNER \citep{adelani-etal-2021-masakhaner} & CoNLL 2003 \citep{tjong-kim-sang-de-meulder-2003-introduction} & mBERT  & Hausa, Igbo, Kinyarwanda, Luganda, Luo, Nigerian-Pidgin, Swahili$^*$, Wolof, Yor\`{u}b\'{a}$^*$ \\
         \cmidrule(lr){2-5}
        Natural Language Inference (NLI) & AmericasNLI \citep{ebrahimi-etal-2022-americasnli} & MultiNLI  \citep{williams-etal-2018-broad} & XLM-R  & Aymara, Ash\'{a}ninka, Bribri, Guarani, N\'{a}huatl, Otom\'{i}, Quechua, Rar\'{a}muri, Shipibo-Konibo, Wixarika \\
        \cmidrule(lr){2-5}
        Question Answering (QA) & XQuAD \citep{artetxe-etal-2020-cross} & SQuAD v1.1 \citep{rajpurkar-etal-2016-squad} & mDeBERTa & Arabic$^\dagger$, Chinese$^\dagger$, German$^\dagger$, Greek$^\dagger$, Hindi$^\dagger$, Romanian$^\dagger$, Russian$^\dagger$, Spanish$^\dagger$, Thai$^\dagger$, Turkish$^\dagger$, Vietnamese$^\dagger$ \\
        \bottomrule
    \end{tabular}}
    \caption{Details of the tasks, datasets, MMTs and languages involved in our zero-shot cross-lingual transfer evaluation. $^*$ denotes low-resource languages and $^\dagger$ high-resource languages seen during MMT pretraining; all other languages are low-resource and unseen. The source language is always English. We \textit{`bistil'} the MMT listed per each task and target language. Further details of all the language and data sources used are provided in Appendix \ref{sec:languages}.}
    \label{tab:tasks}
\end{table*}

\subsection{Stage 2: Task-Specific Distillation} \label{sec:ts-distillation}
After a general bilingual model has been distilled from the teacher MMT in Stage 1, it can be fine-tuned for a specific task. We first obtain the teacher for task-specific distillation by applying task-specific LT-SFT to fine-tune the base MMT (i.e., the teacher in the general distillation phase) for the task in question. This teacher's outputs and representations are then used to fine-tune the bilingual student model, again using task LT-SFT at the student's end. The use of parameter-efficient task adaptation here avoids adding a large number of parameters to the system for each task. The objective during this task-specific fine-tuning consists of the sum of all three losses from \S\ref{sec:distillation-method}: $\loss_\text{attn}$, $\loss_\text{hidden}$, and $\loss_\text{pred}$.


\section{Experimental Setup}
\label{sec:exp}
We largely adopt the evaluation framework of \citet{ansell-etal-2022-composable} for direct comparability with their LT-SFT method, which they apply to undistilled MMTs, and which we apply for task-specific fine-tuning of bilingually distilled MMTs. Specifically, we evaluate zero-shot cross-lingual transfer performance on four representative tasks: dependency parsing, named entity recognition, natural language inference, and QA. While the prior work focused only on low-resource languages, our method is also highly relevant to high-resource languages: the XQuAD QA task \citep{artetxe-etal-2020-cross} provides additional insight into high-resource target language performance. Table~\ref{tab:tasks} summarizes the experimental setup, including the datasets and languages considered in our experiments. In total, we cover a set of 44 typologically and geographically diverse languages, which makes them representative of cross-lingual variation~\citep{ponti-etal-2020-xcopa}.


We experiment with three different MMTs as shown in Table~\ref{tab:tasks}: mBERT~\citep{devlin-etal-2019-bert}, XLM-R$_\text{base}$~\citep{conneau-etal-2020-unsupervised}, and mDeBERTa$_\text{base}$~\cite{he-etal-2021-debertav3}.

\begin{table}[!t]
    \centering
    \def\arraystretch{0.999}
    {\scriptsize
    \resizebox{\linewidth}{!}{
    \begin{tabular}{llcc|cccc}
        \toprule
        \textbf{MMT} & \textbf{Distillation} & \textbf{LRF} & \textbf{DRF} & \textbf{\#L} & \textbf{D} & \textbf{\#V} & \textbf{\#P} \\
        \cmidrule(lr){2-8}
        \multirow{4}{*}{mBERT} & none & - & - & 12 & 768 & 120K & 178M \\
        & \method{D'mBERT} & 2 & - & 6 & 768 & 120K & 135M \\
        & \multirow{2}{*}{\method{BiStil}$^*$} & 2 & - & 6 & 768 & 31K & 67M \\
        & & 3 & - & 4 & 768 & 31K & 53M \\
        \cmidrule(lr){2-8}
        \multirow{3}{*}{XLM-R$_\text{base}$} & none & - & - & 12 & 768 & 250K & 278M \\
        & \multirow{2}{*}{\method{BiStil}$^*$} & 2 & - & 6 & 768 & 28K & 65M \\
        & & 3 & - & 4 & 768 & 28K & 51M \\
        \cmidrule(lr){2-8}
        \multirow{2}{*}{XLM-R$_\text{large}$} & none & - & - & 24 & 1024 & 250K & 560M \\
        & \method{MiniLMv2} & 2 & 2.67 & 12 & 384 & 250K & 118M \\
        \cmidrule(lr){2-8}
        \multirow{3}{*}{mDeBERTa} & none & - & - & 12 & 768 & 250K & 278M \\
        & \multirow{2}{*}{\method{BiStil}$^*$} & 2 & - & 6 & 768 & 41K & 75M \\
        &  & 3 & - & 4 & 768 & 41K & 60M \\
        \bottomrule
    \end{tabular}
    }%
    }%
    \caption{Dimensions of models \textit{before} and \textit{after} distillation. LRF = Layer Reduction Factor; DRF = hidden Dimension Reduction Factor; \#L = number of Transformer Layers; D = hidden Dimension; \#V = Vocabulary size; \#P = total number of model Parameters; \method{D'mBERT} = \method{DistilmBERT}. $^*$ - because of its vocabulary reduction procedure, \method{BiStil} can produce models of slightly different sizes for different languages; the vocabulary sizes and numbers of parameters shown are averages over all \method{BiStil} models trained.}
    \label{tab:model-dimensions}
\end{table}

\subsection{Baselines and Model Variants}
\label{ss:baselines}
We refer to our main method as \method{BiStil}. We compare it with several relevant approaches. First, the \method{LtSft} method \citep{ansell-etal-2022-composable}, a state-of-the-art cross-lingual transfer approach, uses LT-SFT with language adaptation on the base MMT. \method{LtSft} can be seen as an upper bound for \method{BiStil}, allowing us to measure how much the performance suffers as a result of replacing the MMT with its bilingually distilled variant.

For each task except NLI,\footnote{\label{note:distil-no-la} There does not seem to be a multilingually distilled MMT that would provide a suitable comparison on the AmericasNLI task, as MiniLMv2 and other models distilled without an MLM head do not support adaptation to unseen languages through continued pretraining. DistilmBERT on the other hand is based on a generally weaker MMT than XLM-R, which is used as \method{BiStil}'s base MMT for NLI.} we also compare against a multilingually distilled MMT, i.e. with all pretraining languages used for distillation as well. For DP and NER, where mBERT is the base MMT, the distilled MMT is \method{DistilmBERT} \citep{sanh-etal-2019-distilbert}, which is similarly based on mBERT. For QA, where \method{BiStil} uses mDeBERTa as the base MMT, no directly comparable multilingually distilled MMT is available, so we opt for a loose comparison with \method{MiniLMv2} \citep{wang-etal-2021-minilmv2}, distilled from XLM-R$_\text{large}$, which has achieved strong results on cross-lingual transfer in high-resource languages. We perform task-specific fine-tuning with LT-SFT on DistilmBERT and MiniLMv2 in the same way as for the the undistilled MMTs in the \method{LtSft} setting. For DP and NER we also perform language adaptation of DistilmBERT.\footnote{MiniLMv2 does not support language adaptation (see Footnote~\ref{note:distil-no-la}), but this is not as important for the high-resource languages used in XQuAD \citep{ansell-etal-2022-composable}.}

We also consider \method{Scratch}, a setting where we train bilingual models from scratch instead of distilling them from a pretrained MMT. We then apply the same LT-SFT task fine-tuning method as for the other baselines. This comparison allows us to evaluate the benefit of distilling efficient bilingual models from the MMT rather than pretraining the same-sized bilingual models from scratch.

We refer to our main method, with the task-specific distillation stage as described in \S\ref{sec:ts-distillation}, as \method{BiStil-TF} (TF = \textit{teacher forcing}). We also carry out an ablation focused on the second phase of \bistil: here, we consider performing task-specific fine-tuning without the assistance of a teacher, i.e. in the same manner as \method{LtSft}. We refer to this variant as \method{BiStil-ST} (ST =\textit{ self-taught}).

Table~\ref{tab:model-dimensions} provides details of the model sizes, before and after distillation using the above methods, demonstrating the benefits of \bistil with respect to model compactness. 

\subsection{Distillation/Adaptation Training Setup}
We always perform language adaptation of the teacher model during both phases of \bistil and during \method{LtSft} except for mDeBERTa and MiniLMv2\footnote{See Footnote~\ref{note:distil-no-la} for MiniLMv2; mDeBERTa could in theory support language adaptation but its pretraining code was not made publicly available in time to be used in this work.}. For language adaptation of MMTs we use the pretrained language SFTs of~\citet{ansell-etal-2022-composable}, and we train our own for DistilmBERT. Similarly, for the \method{LtSft} baseline, and for task adaptation of the teacher in the \method{BiStil-TF} configuration, we use their pretrained single-source task SFTs or train our own when necessary. When training/distilling our own models or SFTs, we generally choose hyperparameters which match those used to train their SFTs in the original work. See Appendix~\ref{sec:hyperparams} for full training details and hyperparameters of all models in our comparison, and Appendix~\ref{sec:languages} for details of the training corpora.

We experiment with two layer reduction factors (LRF) for \bistil, 2 (a reduction from 12 to 6 layers) and 3 (12 to 4 layers). Whereas the \method{BiStil} setting initializes the model from the teacher (see \S\ref{sec:general-distillation}), the \method{Scratch} setting initializes it randomly.

\begin{table*}[!t]
    \centering
    \def\arraystretch{0.81}
    \resizebox{\linewidth}{!}{
    \begin{tabular}{l|cccccccccccccccc|cc}
	\toprule
	 & \texttt{ar} & \texttt{bm} & \texttt{bxr} & \texttt{fo} & \texttt{gv} & \texttt{hsb} & \texttt{ja} & \texttt{kpv} & \texttt{mt} & \texttt{myv} & \texttt{olo} & \texttt{sa} & \texttt{sme} & \texttt{ug} & \texttt{yue} & \texttt{zh} & \texttt{avg} & \texttt{avg$\Delta$} \\
	\midrule
	\method{LtSft} & 53.6 & 16.5 & 25.9 & 55.5 & 42.4 & 60.5 & 19.7 & 27.2 & 55.4 & 45.3 & 47.8 & 25.2 & 42.1 & 16.7 & 34.0 & 37.0 & 37.8 & - \\
	\hdashline
	\method{DistilmBert} & 47.7 & 9.9 & 19.5 & 49.1 & 31.7 & 53.2 & 16.2 & 20.0 & 43.0 & 34.9 & 37.6 & 17.7 & 31.4 & 11.4 & 28.9 & 33.9 & 30.4 & -7.4 \\
	\method{Scratch, lrf = 2} & 16.9 & 4.9 & 6.7 & 27.8 & 9.1 & 15.2 & 6.7 & 5.6 & 16.1 & 12.7 & 11.1 & 3.5 & 9.3 & 3.9 & 11.5 & 14.6 & 11.0 & -26.8 \\
	\method{BiStil-ST, lrf = 2} & 50.9 & 15.8 & 24.1 & 53.7 & 38.3 & 57.1 & 18.7 & \textbf{23.9} & 52.2 & \textbf{43.7} & \textbf{46.5} & 25.2 & 39.8 & 13.3 & 31.8 & 34.8 & 35.6 & -2.2 \\
	\method{BiStil-ST, lrf = 3} & 48.2 & 16.1 & 23.4 & 52.1 & 35.0 & 55.1 & 18.1 & 22.2 & 49.9 & 40.3 & 41.3 & 22.2 & 37.6 & 13.3 & 30.7 & 33.4 & 33.7 & -4.1 \\
	\method{BiStil-TF, lrf = 2} & \textbf{53.2} & \textbf{16.4} & \textbf{24.6} & \textbf{54.8} & \textbf{39.1} & \textbf{59.0} & \textbf{19.0} & 23.8 & \textbf{54.1} & 43.5 & 46.0 & \textbf{26.9} & \textbf{40.7} & 13.1 & \textbf{32.7} & \textbf{36.4} & \textbf{36.5} & \textbf{-1.3} \\
	\method{BiStil-TF, lrf = 3} & 49.7 & \textbf{16.4} & 24.4 & 52.7 & 36.8 & 57.1 & 18.2 & 21.0 & 52.2 & 41.0 & 43.3 & 25.1 & 38.1 & \textbf{14.5} & 31.3 & 34.9 & 34.8 & -3.0 \\
	\bottomrule
\end{tabular}
    }
    \caption{DP; LAS scores. The results with the smallest gap to the upper-bound \method{LtSft} model are in \textbf{bold}.}
    \label{tab:dp-las}
\end{table*}

\begin{table*}[h]
    \centering
    \def\arraystretch{0.999}
    \footnotesize
    \begin{tabular}{l|ccccccccc|cc}
    	\toprule
    	 & \texttt{hau} & \texttt{ibo} & \texttt{kin} & \texttt{lug} & \texttt{luo} & \texttt{pcm} & \texttt{swa} & \texttt{wol} & \texttt{yor} & \texttt{avg} & \texttt{avg$\Delta$} \\
    	\midrule
    	\method{LtSft} & 83.5 & 76.7 & 67.4 & 67.9 & 54.7 & 74.6 & 79.4 & 66.3 & 74.8 & 71.7 & - \\
    	\hdashline
    	\method{DistilmBert} & 81.1 & 73.2 & 65.3 & 63.4 & 50.0 & 69.2 & 77.7 & 64.4 & 71.2 & 68.4 & -3.3 \\
    	\method{BiStil-ST, lrf = 2} & \textbf{81.3} & 74.1 & 65.9 & 66.7 & 53.5 & 72.1 & 77.1 & 64.6 & 72.8 & 69.8 & -1.9 \\
    	\method{BiStil-ST, lrf = 3} & 80.3 & 74.0 & 63.1 & 64.6 & 54.7 & 69.6 & 76.9 & 68.0 & 70.5 & 69.1 & -2.6 \\
    	\method{BiStil-TF, lrf = 2} & 81.0 & \textbf{74.8} & \textbf{67.5} & \textbf{67.3} & 55.0 & \textbf{72.9} & \textbf{78.4} & \textbf{69.0} & \textbf{75.7} & \textbf{71.3} & \textbf{-0.4} \\
    	\method{BiStil-TF, lrf = 3} & 79.6 & \textbf{74.8} & 64.6 & 64.5 & \textbf{56.7} & 70.6 & 77.2 & 66.1 & 72.8 & 69.6 & -2.1 \\
    	\bottomrule
    \end{tabular}
    \caption{NER; F1 scores.}
    \label{tab:ner}
\end{table*}

\begin{table*}[!t]
    \centering
    \def\arraystretch{0.999}
    \footnotesize
    \resizebox{\linewidth}{!}{
    \begin{tabular}{l|cccccccccc|cc}
    	\toprule
    	 & \texttt{aym} & \texttt{bzd} & \texttt{cni} & \texttt{gn} & \texttt{hch} & \texttt{nah} & \texttt{oto} & \texttt{quy} & \texttt{shp} & \texttt{tar} & \texttt{avg} & \texttt{avg$\Delta$} \\
    	\midrule
    	\method{LtSft} & 58.1 & 44.4 & 47.9 & 63.5 & 42.8 & 52.4 & 48.5 & 62.0 & 50.3 & 43.3 & 51.3 & - \\
    	\hdashline
    	\method{BiStil-TF, lrf = 2} & \textbf{58.9} & \textbf{45.7} & 46.4 & \textbf{62.9} & \textbf{44.3} & 50.8 & \textbf{44.0} & 58.7 & \textbf{47.2} & \textbf{43.1} & \textbf{50.2} & \textbf{-1.1} \\
    	\method{BiStil-TF, lrf = 3} & 57.7 & 43.6 & \textbf{48.1} & 60.9 & 41.3 & \textbf{51.4} & 42.6 & \textbf{59.9} & 45.5 & 40.3 & 49.1 & -2.2 \\
    	\bottomrule
    \end{tabular}
    }
    \caption{NLI accuracy (\%)}
    \label{tab:nli}
\end{table*}

\begin{table*}[!h]
    \centering
    \def\arraystretch{0.999}
    \footnotesize
    \resizebox{\linewidth}{!}{
        \begin{tabular}{l|ccccccccccc|cc}
    	\toprule
    	 & \texttt{ar} & \texttt{de} & \texttt{el} & \texttt{es} & \texttt{hi} & \texttt{ro} & \texttt{ru} & \texttt{th} & \texttt{tr} & \texttt{vi} & \texttt{zh} & \texttt{avg} & \texttt{avg$\Delta$} \\
    	\midrule
    	\method{LtSft} & 56.5 & 64.7 & 61.2 & 62.4 & 57.8 & 69.0 & 61.8 & 56.0 & 56.4 & 57.1 & 60.8 & 60.3 & - \\
    	\hdashline
    	\method{MiniLMv2} & 50.4 & 59.4 & 54.4 & 57.9 & 52.9 & 64.5 & 57.6 & 50.3 & 51.3 & \textbf{53.8} & 55.0 & 55.2 & -5.1 \\
    	\method{BiStil-TF, lrf = 2} & \textbf{53.5} & \textbf{62.2} & \textbf{55.4} & \textbf{59.8} & \textbf{54.5} & \textbf{66.2} & \textbf{58.3} & \textbf{54.4} & \textbf{53.1} & 53.4 & \textbf{55.7} & \textbf{57.0} & \textbf{-3.4} \\
        \method{BiStil-TF, lrf = 3} & 44.3 & 55.0 & 44.1 & 55.2 & 46.1 & 59.5 & 51.3 & 42.4 & 48.3 & 44.6 & 50.9 & 49.3 & -11.1 \\
    	\bottomrule
    \end{tabular}
    }
    \caption{XQuAD; Exact Match scores.}
    \label{tab:qa-em}
\end{table*}

\section{Results and Discussion}
The results in terms of task performance are summarized in Tables \ref{tab:dp-las}-\ref{tab:qa-em}. As expected, \method{LtSft} on the undistilled MMTs performs best across all tasks. However, \method{BiStil}-TF with reduction factor 2 is not much worse, with a degradation in performance not exceeding 1.3 points relative to \method{LtSft} on DP, NER and NLI. The larger gap of 3.4 EM points on QA is likely a result of the fact that the base MMT is much more thoroughly pretrained on the high-resource languages found in XQuAD than on the lower-resource languages found in the datasets for the other tasks. It is therefore harder for \method{BiDistil} to achieve the base MMT's depth of knowledge of the target language during its relatively short distillation training time. \method{BiStil-TF, lrf = 2} nevertheless outperforms MiniLMv2 on QA by 1.7 EM points, despite MiniLMv2 receiving 320 times more training than each \method{BiDistil} model, or roughly 6 times more per language\footnote{MiniLMv2 is trained for 1M steps with a batch size of 256 and max sequence length of 512; \method{BiDistil} for 200K steps with a batch size of 8 and max sequence length of 256.}. 

Furthermore, \method{BiStil-TF, lrf = 2} significantly outperforms \method{DistilmBERT}, with a 6.1 LAS gap on DP and 2.9 F1 gap on NER. \method{BiStil, lrf = 2} produces models roughly half the size of \method{DistilmBERT} and that, once again, are trained for vastly less time\footnote{\citet{sanh-etal-2019-distilbert} note that their monolingual DistilBERT model was trained on 8 16GB V100 GPUs for approximately 90 hours. Our \method{BiStil} models are trained on a single 10GB RTX 3080 GPU for approximately 9 hours.}.


Training bilingual models from \textsc{Scratch} performs poorly, lagging behind the other methods by more than 20 points on DP.\footnote{As this method is clearly inferior, we opted to reduce computational expense by not repeating it for other tasks.} One crucial weakness of \textsc{Scratch}, besides its reduced monolingual performance, is a lack of alignment between its representations of the source and target languages, severely impairing cross-lingual transfer. This highlights the advantage of distilling a bilingual model from an MMT within which cross-lingual alignment is already present.

Interestingly, when we evaluate the \method{Scratch} models on their \textit{English} DP performance, we obtain an average UAS/LAS score of 81.8/77.1, which is much more competitive in relative terms with the \method{BiStil-TF, lrf = 2} English DP score of 91.0/88.2 than the corresponding comparison in average target language DP scores of 29.9/11.0 to 55.5/36.5. This suggests that an even larger factor in \method{Scratch}'s weakness than its poor monolingual performance is a lack of alignment between its representations of the source and target languages, severely impairing cross-lingual transfer. This highlights the advantage of distilling a bilingual model from an MMT within which cross-lingual alignment is already present.

As expected, the performance of \method{BiStil} is somewhat weaker with a larger layer reduction factor of 3, though this is heavily task-dependent. With an LRF of 3, \method{BiStil-TF} still comfortably outperforms \method{DistilmBERT} on DP and NER, and does not fall much behind LRF = 2 for NLI. However, we observe a considerable degradation in performance for LRF = 3 for QA; this may indicate that a 4-layer Transformer struggles to adapt to this particular task, or that for this architecture the modest training time is not sufficient to approach the base MMT's understanding of the source and target languages. 

\begin{table}[!t]
    \centering
    \begin{subtable}[!t]{0.8\linewidth}
        \centering
        \def\arraystretch{0.999}
        \footnotesize
        \begin{tabular}{l|cc|c}
        	\toprule
        	 & \texttt{cpu $\uparrow$} & \texttt{gpu $\uparrow$} & \texttt{flops $\downarrow$} \\
        	\midrule
        	\method{DistilmBERT} & 1.41x & 1.03x & 0.61x \\
        	\method{BiStil, lrf = 2} & 1.44x & 1.25x & 0.61x \\
        	\method{BiStil, lrf = 3} & 1.71x & 1.36x & 0.48x \\
        	\bottomrule
        \end{tabular}
        \caption{DP efficiency}
        \label{tab:dp-efficiency}
    \end{subtable}
    \begin{subtable}[!t]{0.8\linewidth}
        \centering
        \def\arraystretch{0.999}
        \footnotesize
        \begin{tabular}{l|cc|c}
        	\toprule
        	 & \texttt{cpu $\uparrow$} & \texttt{gpu $\uparrow$} & \texttt{flops $\downarrow$} \\
        	\midrule
        	\method{DistilmBERT} & 1.93x & 1.94x & 0.50x \\
        	\method{BiStil, lrf = 2} & 1.97x & 1.98x & 0.50x \\
        	\method{BiStil, lrf = 3} & 2.97x & 2.78x & 0.33x \\
        	\bottomrule
        \end{tabular}
        \caption{NER efficiency}
        \label{tab:ner-efficiency}
    \end{subtable}

    \begin{subtable}[!t]{0.8\linewidth}
        \centering
        \def\arraystretch{0.999}
        \footnotesize
        \begin{tabular}{l|cc|c}
        	\toprule
        	 & \texttt{cpu $\uparrow$} & \texttt{gpu $\uparrow$} & \texttt{flops $\downarrow$} \\
        	\midrule
        	\method{BiStil, lrf = 2} & 2.02x & 1.97x & 0.50x \\
        	\method{BiStil, lrf = 3} & 2.89x & 2.85x & 0.33x \\
        	\bottomrule
        \end{tabular}
        \caption{NLI efficiency}
        \label{tab:nli-efficiency}
    \end{subtable}
    \begin{subtable}[!t]{0.8\linewidth}
        \centering
        \def\arraystretch{0.999}
        \footnotesize
        \begin{tabular}{l|cc|c}
        	\toprule
        	 & \texttt{cpu $\uparrow$} & \texttt{gpu $\uparrow$} & \texttt{flops $\downarrow$} \\
        	\midrule
        	\method{MiniLMv2} & 4.25x & 3.44x & 0.21x \\
        	\method{BiStil, lrf = 2} & 1.99x & 1.85x & 0.50x \\
            \method{BiStil, lrf = 3} & 2.85x & 2.42x & 0.33x \\
        	\bottomrule
        \end{tabular}
        \caption{QA efficiency}
        \label{tab:qa-efficiency}
    \end{subtable}
    \caption{Relative inference speed and FLOP cost. Values are given relative to \method{LtSft} without distillation, i.e. a speed reading of ``2.00x'' means the distilled model can on average process twice as many inference examples per second as the undistilled MMT. Likewise a FLOPs reading of ``0.50x'' would mean that the distilled model on average requires half as many FLOPs to process an inference example as the undistilled MMT does.}
    \label{tab:efficiency}
\end{table}

Table~\ref{tab:efficiency} presents an analysis of the inference time efficiency. We measure the inference speed both on CPU with batch size 1 and GPU with the same batch size as during task-specific training. We also calculate the number of floating-point operations (FLOPs) per example using \href{https://github.com/facebookresearch/fvcore}{fvcore}, measured during an inference pass over the test set of the first language in each task.




For NER, NLI and QA, the efficiency results conform quite closely to the intuitive expectation that a model's inference time should scale linearly with its number of layers; that is, \method{BiDistil} with LRF = 2 is generally around twice as fast as the base MMT. For DP, we observe a seemingly sub-linear scaling which is caused by the very large biaffine parsing head, consisting of $\sim$23M parameters. The significant cost of applying the model head contributes equally to all models regardless of their degree of distillation. Despite having a moderate LRF of 2, \method{MiniLMv2} exhibits impressive speed as a result of the fact that it additionally has a smaller hidden dimension than its teacher (see Table~\ref{tab:model-dimensions}), a technique which we do not consider for \method{BiDistil}, but may be a promising avenue for future work.

We argue that \method{BiDistil} accomplishes its aim by achieving two- to three-fold reductions in inference time and model size without sacrificing much in the way of raw performance. Its superior performance relative to multilingually distilled models despite its comparatively very modest training budget supports the assertion that specializing multilingual models for a specific transfer pair during distillation helps to avoid performance degradation resulting from the curse of multilinguality.

\section{Related Work}

One strand of prior work focuses on parameter-efficient adaptation of pretrained MMTs, i.e. adaptation by adding/modifying a small subset of parameters. Adapters~\citep{rebuffi-etal-2017-learning,houlsby-etal-2019-parameter} have been used extensively for this purpose~\citep{ustun-etal-2020-udapter}, with the MAD-X framework of~\citet{pfeiffer-etal-2020-mad} becoming a starting point for several further developments~\citep{vidoni-etal-2020-orthogonal,wang-etal-2021-efficient-test,parovic-etal-2022-bad}, where a notable theme is adapting MMTs to unseen languages~\citep{ansell-etal-2021-mad-g,pfeiffer-etal-2021-unks}. \citet{ansell-etal-2022-composable} propose composable sparse fine-tunings as an alternative to adapters. 

\citet{pfeiffer-etal-2022-lifting} create a modular MMT from scratch, where some parameters are shared among all languages and others are language-specific. This allows the model to dedicate considerable capacity to every language without each language-specific model becoming overly large; thus it is quite similar in its aims to this work.

A variety of approaches have been proposed for general distillation of pretrained language models. The simplest form uses only soft target probabilities predicted by the teacher model as the training signal for the student \citep{sanh-etal-2019-distilbert}. Other approaches try to align the hidden states and self-attention distributions of the student and teacher~\citep{sun-etal-2020-mobilebert,jiao-etal-2020-tinybert} and/or finer-grained aspects of the self-attention mechanism~\citep{wang-etal-2020-minilm,wang-etal-2021-minilmv2}. \citet{mukherjee-etal-2021-xtreme} initialize the student's embedding matrix with a factorization of the teacher's for better performance when their hidden dimensions differ. Of these, \citet{sanh-etal-2019-distilbert,wang-etal-2020-minilm,wang-etal-2021-minilmv2,mukherjee-etal-2021-xtreme} apply their methods to produce distilled versions of MMTs.

\citet{parovic-etal-2022-bad} adapt pretrained MMTs to specific transfer pairs with adapters; this approach is similar to ours in spirit, but it is aimed towards improving performance rather than efficiency. \citet{minixhofer-etal-2022-wechsel} learn to transfer full monolingual models across languages. The only work prior we are aware of which creates purely bilingual models for cross-lingual transfer is that of \citet{tran-2020-english}. This approach starts with a monolingual pretrained source language model, initializes target language embeddings via an alignment procedure, and then continues training the model with the added target embeddings on both languages.

\section{Conclusions}
While MMTs are an effective tool for cross-lingual transfer, their broad language coverage makes them unnecessarily costly to deploy in the frequently-encountered situation where capability is required in only a single, often low-resource, language. We have proposed \bistil, a method of training more efficient models suited to this scenario which works by distilling an MMT using only the source-target language pair of interest. We show that this approach produces models that offer an excellent trade off between target language performance, efficiency, and model compactness. The `bistilled' models exhibit only a slight decrease in performance relative to their base MMTs whilst achieving considerable reduction in both model size and inference time. Their results also compare favorably to those of multilingually distilled MMTs despite receiving substantially less training even on a per-language basis. 


\section*{Limitations}
While the results of our experiments seem sufficient to validate the concept and our general approach to bilingual distillation, we have not carried out a detailed systematic analysis of alternative implementations of the various aspects of our methods, such as different student model initializations, distillation objectives and hyperparameter settings. Furthermore, our \method{BiStil} models are likely undertrained due to limited computational resources. Consequently, we do not claim our specific implementation of bilingual distillation to be optimal or even close to optimal. Areas that warrant further investigation toward realizing the full potential of this approach include the use of hidden dimension reduction, which yielded impressive speed gains for MiniLMv2 in our experiments, and other innovations in distillation such as progressive knowledge transfer \citep{mukherjee-etal-2021-xtreme}.

With the exception of improved efficiency, our \method{BiStil} models inherit the limitations of the MMTs from which they are distilled; notably, there is a discrepancy between the performance on high- and low-resource languages resulting from the distribution of data used during MMT pretraining.

In this work, we have only considered English as the source language; some target languages may benefit from other transfer sources. Future work may also consider the use of multi-source transfer, which would entail distilling with more than two languages. Here the challenge would be optimizing the balance of model capacity allocated to source languages versus the target language. 

\section*{Acknowledgements}
Alan wishes to thank David and Claudia Harding for their generous support via the Harding Distinguished Postgraduate Scholarship Programme. Ivan Vuli\'{c} is supported by a personal Royal Society University Research Fellowship \textit{`Inclusive and Sustainable Language Technology for a Truly Multilingual World'} (no 221137; 2022--).

\bibliography{anthology,custom}
\bibliographystyle{acl_natbib}

\appendix

\section{Training Details and Hyperparameters} \label{sec:hyperparams}
As we evaluate over many languages and tasks, we carry out a single run per (task, language, configuration) triple.

\subsection{Language Distillation/Adaptation} 
The following are constant across all language distillation/SFT training: we use a batch size of 8 and a maximum sequence length of 256; model checkpoints are evaluated every 1,000 steps (5,000 for high-resource languages) on a held-out set of 5\% of the corpus (1\% for high-resource languages), and the one with the smallest loss is selected at the end of training; we use the AdamW optimizer \citep{loshchilov-hutter-2019-decoupled} with linear decay without any warm-up. 

During LT-SFT training of DistilmBERT's language SFTs, the dense and sparse fine-tuning phases each last the lesser of 100,000 steps or 200 epochs, but at least 30,000 steps if 200 epochs is less. The initial learning rate is $5 \cdot 10^{-5}$. The SFT density is set to 4\%.\footnote{This is similar but not identical to the density used by \citet{ansell-etal-2022-composable}, who use a very specific number of trainable parameters for comparability to their baseline; we prefer to use a round number.}

When distilling bilingual models or learning them from scratch, training lasts 200,000 steps (to equal the total length of the two phases of LT-SFT training). The initial learning rate is $10^{-4}$. The model architecture and hyperparameters are identical to the teacher MMT's other than a reduction in the number of layers and the use of vocabulary reduction as described in \S\ref{sec:general-distillation}.

\subsection{Task Distillation/Adaptation} 
For DP and NER, we train task SFTs for 3 epochs in the dense phase of LT-SFT and 10 epochs in the sparse phase, evaluating the model checkpoint on the validation set at the end of each epoch, and taking the best checkpoint at the end of training. The selection metric is labeled attachment score for DP and F1-score for NER. The initial learning rate is $5 \cdot 10^{-5}$ with linear decay.  For NER, we use the standard token-level single-layer multi-class model head. For DP, we use the shallow variant \citep{glavas-vulic-2021-supervised} of the biaffine dependency parser of \citet{dozat-manning-2017-deep}. For NLI, we train for 5 epochs with batch size 32, with checkpoint evaluation on the validation set every 625 steps, and an initial learning rate of $2 \cdot 10^{-5}$. We apply a two-layer multi-class classification head atop the model output corresponding to the \texttt{[CLS]} token. For QA, we train for 5 epochs with a batch size of 12, with checkpoint evaluation every 2000 steps and an initial learning rate of $3 \cdot 10^{-5}$. The single-layer model head independently predicts the start and end positions of the answer span, and at inference time the span whose endpoints have the largest sum of logits is selected.

We set the density of our task SFTs to 8\%, which \citet{ansell-etal-2022-composable} found to offer the best task performance in all their experiments.

\onecolumn
\section{Languages} \label{sec:languages}

\begin{table*}[!h]
    \centering
    \scriptsize
    \resizebox{\linewidth}{!}{
    \begin{tabular}{m{0.1\textwidth} | m{0.15\textwidth} | m{0.1\textwidth} | m{0.2\textwidth} | m{0.15\textwidth}| m{0.30\textwidth}}
        \toprule
        \textbf{Task} & \textbf{Language} & \textbf{ISO Code} & \textbf{Family} & \textbf{UD Treebank} & \textbf{Corpus source(s)} \\
        \midrule
        Source & English & en & Indo-European, Germanic & EWT & Wikipedia \\
        \midrule
        \multirow{17}{*}{DP} & Arabic & ar & Afro-Asiatic, Semitic & PUD & \multirow{16}{*}{Wikipedia} \\
        & Bambara & bm & Mande & CRB & \\
        & Buryat & bxr & Mongolic & BDT & \\
        & Cantonese & yue & Sino-Tibetan & HK & \\
        & Chinese & zh & Sino-Tibetan & GSD & \\
        & Erzya & myv & Uralic, Mordvin & JR & \\
        & Faroese & fo & Indo-European, Germanic & FarPaHC & \\
        & Japanese & ja & Japanese & GSD & \\
        & Livvi & olo & Uralic, Finnic & KKPP & \\
        & Maltese & mt & Afro-Asiatic, Semitic & MUDT & \\
        & Manx & gv & Indo-European, Celtic & Cadhan & \\
        & North Sami & sme & Uralic, Sami & Giella & \\
        & Komi Zyrian & kpv & Uralic, Permic & Lattice & \\
        & Sanskrit & sa & Indo-European, Indic & UFAL & \\
        & Upper Sorbian & hsb & Indo-European, Slavic & UFAL & \\
        & Uyghur & ug & Turkic, Southeastern & UDT & \\
        \midrule
        \multirow{9}{*}{NER} & Hausa & hau & Afro-Asiatic, Chadic & \multirow{9}{*}{N/A} & Wikipedia \\
        & Igbo & ibo & Niger-Congo, Volta-Niger & & Wikipedia \\
        & Kinyarwanda & kin & Niger-Congo, Bantu & & Wikipedia \\
        & Luganda & lug & Niger-Congo, Bantu & & Wikipedia \\
        & Luo & luo & Nilo-Saharan & & \href{https://github.com/Pogayo/Luo-News-Dataset}{Luo News Dataset} \citep{adelani-etal-2021-masakhaner} \\
        & Nigerian-Pidgin & pcm & English Creole & & JW300 \citep{agic-vulic-2019-jw300} \\
        & Swahili & swa & Niger-Congo, Bantu & & Wikipedia \\
        & Wolof & wol & Niger-Congo, Senegambian & & Wikipedia \\
        & Yor\`{u}b\'{a} & yor & Niger-Congo, Volta-Niger & & Wikipedia \\
        \midrule
        \multirow{10}{*}{NLI} & Aymara & aym & Aymaran & \multirow{10}{*}{N/A} & \citet{tiedemann-2012-parallel}; Wikipedia \\
        & Ash\'aninka & cni & Arawakan & & \citet{ortega-etal-2020-overcoming,cushimariano:prel:08,mihas:anaani:11,bustamante-etal-2020-data} \\
        & Bribri & bzd & Chibchan, Talamanca & & \citet{feldman-coto-solano-2020-neural} \\
        & Guarani & gn & Tupian, Tupi-Guarani & & \citet{chiruzzo-etal-2020-development}; Wikipedia \\
        & N\'ahuatl & nah & Uto-Aztecan, Aztecan & & \citet{gutierrez-vasques-etal-2016-axolotl}; Wikipedia \\
        & Otom\'i & oto & Oto-Manguean, Otomian & & \href{https://tsunkua.elotl.mx/about/}{H\~{n}\"{a}h\~{n}u Online Corpus} \\
        & Quechua & quy & Quechuan & & \citet{agic-vulic-2019-jw300}; Wikipedia \\
        & Rar\'amuri & tar & Uto-Aztecan, Tarahumaran & & \citet{brambila-1976-diccionario} \\
        & Shipibo-Konibo & shp & Panoan & & \citet{galarreta-etal-2017-corpus,bustamante-etal-2020-data} \\
        & Wixarika & hch & Uto-Aztecan, Corachol & & \citet{mager2018probabilistic} \\
        \midrule
        \multirow{11}{*}{QA} & Arabic & ar & Afro-Asiatic, Semitic & \multirow{11}{*}{N/A} & \multirow{11}{*}{Wikipedia} \\
        & Chinese & zh & Sino-Tibetan & & \\
        & German & de & Indo-European, Germanic & & \\
        & Greek & el & Indo-European, Greek & \\
        & Hindi & hi & Indo-European, Indic & \\
        & Romanian & ro & Indo-European, Romance & & \\
        & Russian & ru & Indo-European, Slavic & & \\
        & Thai & th & Tai-Kadai, Kam-Tai & & \\
        & Turkish & tr & Turkic, Southwestern & & \\
        & Vietnamese & vi & Austro-Asiatic, Viet-Muong & \\
        \bottomrule
    \end{tabular}
    }
    \caption{Details of the languages and data used for training and evaluation of SFTs and adapters. The corpora of \citet{bustamante-etal-2020-data} are available at \url{https://github.com/iapucp/multilingual-data-peru}; all other NLI corpora mentioned are available at \url{https://github.com/AmericasNLP/americasnlp2021}.}
    \label{tab:languages}
\end{table*}

\section{Additional Results} \label{sec:additional-results}

\begin{table*}[!h]
    \centering
    \resizebox{\linewidth}{!}{
    \begin{tabular}{l|cccccccccccccccc|cc}
    	\toprule
    	 & \texttt{ar} & \texttt{bm} & \texttt{bxr} & \texttt{fo} & \texttt{gv} & \texttt{hsb} & \texttt{ja} & \texttt{kpv} & \texttt{mt} & \texttt{myv} & \texttt{olo} & \texttt{sa} & \texttt{sme} & \texttt{ug} & \texttt{yue} & \texttt{zh} & \texttt{avg} & \texttt{avg$\Delta$} \\
    	\midrule
    	\method{LtSft} & 70.8 & 43.1 & 49.2 & 68.2 & 60.0 & 73.7 & 36.9 & 50.5 & 74.6 & 65.9 & 66.4 & 49.5 & 58.0 & 36.4 & 51.1 & 59.8 & 57.1 & - \\
    	\midrule
    	\method{DistilmBert} & 65.7 & 34.4 & 42.3 & 63.0 & 52.8 & 67.6 & 32.1 & 42.2 & 65.4 & 58.6 & 59.6 & 44.1 & 51.2 & 29.2 & 47.0 & 56.1 & 50.7 & -6.4 \\
    	\method{Scratch, lrf = 2} & 38.5 & 26.6 & 24.8 & 44.9 & 35.4 & 33.5 & 18.6 & 23.4 & 42.9 & 31.5 & 30.2 & 23.0 & 26.1 & 12.3 & 30.8 & 35.6 & 29.9 & -27.2 \\
    	\method{BiStil-ST, lrf = 2} & 68.0 & 41.6 & 45.7 & 66.3 & 56.6 & 70.9 & 34.1 & \textbf{48.2} & 71.0 & \textbf{64.5} & \textbf{64.3} & 48.9 & \textbf{57.6} & \textbf{34.5} & 49.4 & 56.7 & 54.9 & -2.2 \\
    	\method{BiStil-ST, lrf = 3} & 65.5 & 42.5 & 45.9 & 64.1 & 52.7 & 68.1 & 33.2 & 46.5 & 68.0 & 62.0 & 61.5 & 46.9 & 55.1 & 32.4 & 48.6 & 55.3 & 53.0 & -4.1 \\
    	\method{BiStil-TF, lrf = 2} & \textbf{70.3} & 43.4 & 46.8 & \textbf{67.1} & \textbf{57.7} & \textbf{72.4} & \textbf{34.5} & 47.6 & \textbf{72.7} & 64.2 & 62.6 & \textbf{50.5} & 57.4 & 32.3 & \textbf{49.8} & \textbf{58.6} & \textbf{55.5} & \textbf{-1.6} \\
    	\method{BiStil-TF, lrf = 3} & 67.0 & \textbf{43.9} & \textbf{47.6} & 65.1 & 54.2 & 70.0 & 33.4 & 44.2 & 69.7 & 62.3 & 61.8 & 49.2 & 55.1 & 33.3 & 48.9 & 56.5 & 53.9 & -3.3 \\
    	\bottomrule
    \end{tabular}
    }
    \caption{DP UAS score}
    \label{tab:dp-uas}
\end{table*}

\begin{table*}[!h]
    \centering
    \footnotesize
    \resizebox{\linewidth}{!}{
    \begin{tabular}{l|ccccccccccc|cc}
    	\toprule
    	 & \texttt{ar} & \texttt{de} & \texttt{el} & \texttt{es} & \texttt{hi} & \texttt{ro} & \texttt{ru} & \texttt{th} & \texttt{tr} & \texttt{vi} & \texttt{zh} & \texttt{avg} & \texttt{avg$\Delta$} \\
    	\midrule
    	\method{LtSft} & 73.0 & 80.5 & 78.6 & 80.6 & 74.3 & 82.4 & 77.8 & 69.7 & 72.2 & 76.5 & 68.9 & 75.9 & - \\
    	\midrule
    	\method{MiniLMv2} & 66.4 & 75.5 & 72.4 & 76.6 & 69.6 & 78.3 & 74.0 & 63.8 & 67.6 & \textbf{73.3} & \textbf{64.6} & 71.1 & -4.8 \\
    	\method{BiStil-TF, lrf = 2} & \textbf{69.4} & \textbf{77.4} & \textbf{73.8} & \textbf{77.6} & \textbf{69.7} & \textbf{79.1} & \textbf{75.0} & \textbf{66.7} & \textbf{68.8} & 72.8 & 64.5 & \textbf{72.3} & \textbf{-3.6} \\
        \method{BiStil-TF, lrf = 3} & 62.4 & 70.7 & 63.3 & 74.7 & 61.4 & 73.4 & 68.7 & 54.3 & 62.9 & 63.0 & 60.4 & 65.0 & -10.9 \\
    	\bottomrule
    \end{tabular}
    }
    \caption{XQuAD F1 score}
    \label{tab:qa-f1}
\end{table*}

\end{document}